\documentclass[10pt,twocolumn,letterpaper]{article}

\usepackage[utf8]{inputenc}
\usepackage[T1]{fontenc}
\usepackage{lmodern}
\usepackage{amsmath,amssymb}

\usepackage{graphicx}
\usepackage{booktabs}
\usepackage{tabularx}
\usepackage[table]{xcolor} 
\usepackage{colortbl}      

\usepackage{algorithm}
\usepackage{algpseudocode} 

\usepackage[pagebackref,breaklinks,colorlinks,allcolors=blue]{hyperref}
\usepackage[nameinlink,noabbrev]{cleveref}

\usepackage{comment}
\usepackage{graphicx}
\usepackage{amsmath,amssymb}
\usepackage{booktabs}
\usepackage{tabularx}
\usepackage{caption}
\usepackage{subcaption}

\usepackage{xcolor}

\usepackage[pagebackref,breaklinks,colorlinks,allcolors=blue]{hyperref}
\usepackage[nameinlink,noabbrev]{cleveref}

\usepackage{todonotes}

\usepackage[margin=1in]{geometry}

\title{CLASP: Adaptive Spectral Clustering for Unsupervised Per-Image Segmentation}

\author{
  Max Curie\\
  Integral Ad Science, New York, USA\\
  \texttt{txing@integralads.com}
  \and
  Paulo da Costa\\
  Integral Ad Science, New York, USA\\
  \texttt{pdacosta@integralads.com}
}

\date{} 

\begin{document}
\maketitle

\begin{abstract}
We introduce CLASP (Clustering via Adaptive Spectral Processing), a lightweight framework for unsupervised image segmentation that operates without any labeled data or finetuning. CLASP first extracts per patch features using a self supervised ViT encoder (DINO); then, it builds an affinity matrix and applies spectral clustering. To avoid manual tuning, we select the segment count automatically with a eigengap silhouette search, and we sharpen the boundaries with a fully connected DenseCRF. Despite its simplicity and training free nature, CLASP attains competitive mIoU and pixel accuracy on COCO Stuff and ADE20K, matching recent unsupervised baselines. The zero training design makes CLASP a strong, easily reproducible baseline for large unannotated corpora especially common in digital advertising and marketing workflows such as brand safety screening, creative asset curation, and social media content moderation.
\end{abstract}
\section{Introduction}

\begin{figure*}[H]
    \centering
    \includegraphics[width=0.9\textwidth]{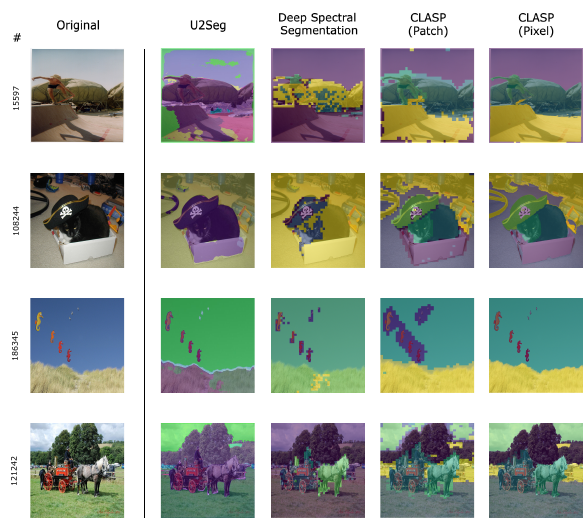}
    \caption{Comparative analysis of image segmentation methods on the COCO-Stuff dataset. The first column shows the original images, while subsequent columns present segmentation outputs from U2Seg, Deep Spectral Segmentation, and our proposed method, \textbf{CLASP}, in both its \textbf{Patch-based} and \textbf{Pixel-based} variants. Despite its simplicity, CLASP produces structurally consistent and visually interpretable segmentations that can serve as fast, training-free region proposals — particularly useful in large-scale multimedia workflows where semantic alignment is not a requirement.}
    \label{fig:segmentation_comparison}
\end{figure*}
Unsupervised image segmentation—the task of discovering coherent regions without human labels—is increasingly important as visual data scale beyond the limits of manual annotation. In domains such as digital advertising and marketing, social media platforms generate millions of photos and videos daily. These assets must be automatically processed for tasks like brand-safety screening, product-catalog tagging, and content filtering, all of which benefit from fast, label-free segmentation. Similar demands arise in autonomous driving, remote sensing, and medical imaging \cite{long2015fully,ronneberger2015u,cordts2016cityscapes,maggiori2017high,krizhevsky2012imagenet}.

While supervised methods achieve strong performance \cite{cheng2022mask2former,xie2021segformer,long2015fully,ronneberger2015u,chen2018encoder,zhao2017pyramid}, they require large-scale pixel-level annotations that are expensive to collect. Self-supervised vision transformers such as DINO \cite{caron2021emerging} offer an alternative, producing patch-level embeddings that encode semantic structure without labels. Several recent approaches apply spectral clustering to these features \cite{wang2022tokencut,wang2023cut}, but often require a fixed number of clusters or rely on iterative algorithms like $k$-means.

We address \emph{unsupervised per-image segmentation}—inferring region-level structure within individual images, without requiring class-consistent clusters across a dataset. This task matches practical needs in large-scale multimedia classification, where the goal is to identify meaningful regions (e.g., objects, people, text zones) in each image independently.

We propose \textbf{CLASP} (\textbf{CL}ustering via \textbf{A}daptive \textbf{S}pectral \textbf{P}rocessing), a training-free segmentation framework that combines DINO ViT embeddings with adaptive spectral clustering. CLASP builds a cosine-affinity graph over patch features, computes its normalised Laplacian, and selects the number of clusters using an \emph{eigengap–silhouette} heuristic that balances global structure with local compactness. A single pass over the spectral embedding assigns segment labels; an optional DenseCRF \cite{krahenbuehl2011efficient} sharpens the final masks. CLASP exposes only one hyperparameter (graph bandwidth), and requires no fine-tuning or supervision.

We evaluate CLASP on COCO-Stuff and ADE20K, achieving 36\% mIoU and 64\% pixel accuracy on COCO-Stuff—competitive with more complex pipelines while remaining significantly simpler.

Our contributions are:
\begin{itemize}
    \item A fully training-free framework for unsupervised per-image segmentation that leverages self-supervised ViT features.
    \item An adaptive cluster-count selection strategy using eigengap and silhouette scores, removing the need for manual tuning.
    \item A lightweight, reproducible baseline suited for large-scale, label-free region discovery.
\end{itemize}

    \section{Literature Review}
\label{sec:lit_review}

Unsupervised segmentation has gained momentum as powerful self-supervised vision transformers, particularly DINO~\cite{caron2021emerging}, provide high-quality feature representations without requiring annotations. Early approaches such as IIC~\cite{ji2019invariant} and PiCIE~\cite{cho2021picie} introduced contrastive learning objectives to cluster pixels into semantic groups, enabling models to learn spatial structure from scratch. These methods, however, typically involve training a segmentation head and require careful tuning of loss functions and training schedules.

TokenCut~\cite{wang2022tokencut} marked a shift toward leveraging pretrained features by constructing a fully connected graph over ViT patch embeddings and applying Normalized Cuts to extract salient foreground masks for single-object segmentation. MaskCut~\cite{wang2024unsam} and CutLER~\cite{wang2023cutler} extended this approach to multi-object segmentation through recursive cuts and student-teacher distillation. These methods inspired a class of ``deep spectral'' pipelines that process each image independently using graph-based clustering. However, they often require a pre-defined number of clusters or multiple passes, which can limit scalability and robustness.

To improve semantic consistency across images, methods like STEGO~\cite{hamilton2022unsupervised} and U2Seg~\cite{niu2023unsupervised} convert initial clusters into pseudo-labels and distill them into trainable segmentation networks. STEGO learns patch-level correspondence through contrastive loss, while U2Seg uses self-training to refine clusters and segment both objects and instances. These pipelines often achieve strong performance, but they introduce supervision-like mechanisms, require multi-stage training, and add sensitivity to hyperparameter choices.

Recent work explores segment hierarchy and adaptive granularity. ACSeg~\cite{li2023acseg} dynamically chooses the number of clusters per image by optimizing intra- and inter-cluster contrast, while HAUS~\cite{rossetti2024hierarchy} introduces hierarchy-agnostic segmentation using probabilistic region proposals. DGS~\cite{sick2024unsupervised} improves spatial consistency by incorporating monocular depth maps into the segmentation pipeline. While methods like HAUS~\cite{rossetti2024hierarchy} aim to model semantic structure hierarchically across images, CLASP targets a simpler and more focused goal: per-image segmentation using adaptive spectral clustering. CLASP makes no assumptions about inter-image consistency or semantic label propagation, making it ideal for training-free applications where individual image structure is more relevant than dataset-level coherence.

Alternative paradigms have emerged. DiffuseSeg~\cite{tian2024diffuse} leverages self-attention maps from pretrained diffusion models to produce semantic regions, while UnSAM~\cite{wang2024unsam} constructs a promptable segmentation model by refining cluster assignments with additional architectural supervision. However, both methods assume a dataset-level notion of class consistency—i.e., the same cluster ID corresponds to semantically similar regions across images—which requires global alignment of features or explicit prompts. In contrast, our method performs per-image segmentation without assuming inter-image label consistency. 

Most related to our work are per-image deep spectral methods such as Deep Spectral Segmentation~\cite{melas2022deep}, which perform clustering on patch embeddings via graph Laplacians but typically require fixed cluster counts or iterative clustering like $k$-means. CLASP builds on this tradition by introducing a lightweight, adaptive mechanism for selecting the number of clusters via eigengap and silhouette heuristics. The method produces segmentation masks in a single forward pass—without any training, fine-tuning, or distillation—offering a reproducible, plug-and-play alternative for large-scale region discovery tasks.

\begin{figure*}[t]
\begin{center}
\includegraphics[width=0.8\linewidth]{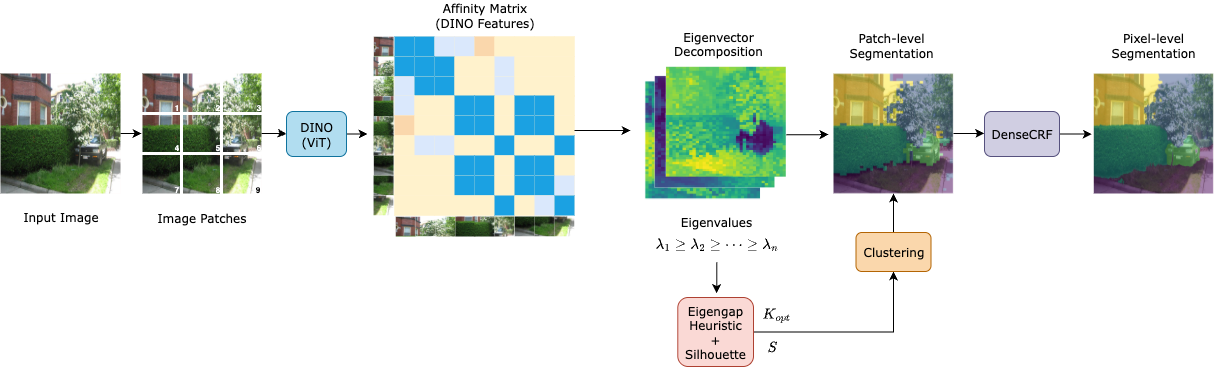}
\end{center}
\caption{Illustrated workflow for semantic segmentation using CLASP. The process begins with image patch extraction, followed by feature encoding using a DINO (ViT) model. An affinity matrix is then constructed based on the extracted features, which undergoes eigenvector decomposition. The eigengap heuristic and silhouette analysis determine the number of clusters, guiding the spectral clustering process. The resulting patch-level segmentation is further refined using a DenseCRF, producing a final pixel-level segmentation output.}
\label{fig:workflow}
\end{figure*}

\subsection{CLASP (Our Method)}
Our proposed method, CLASP, builds on the deep spectral clustering framework while directly addressing the limitations of previous approaches. Like earlier spectral methods, CLASP begins by constructing an affinity graph from self-supervised features and performing spectral decomposition. In contrast to previous methods, however, CLASP automatically determines the number of meaningful clusters by analyzing the eigenvalue spectrum using an eigengap heuristic. This adaptive strategy eliminates the need for a pre-specified $K$, thereby removing the dependence on external clustering steps such as $k$-means and avoiding their associated randomness and instability.

By directly deriving a discrete partition from the selected eigenvectors, CLASP generates a fully defined segmentation mask in a single forward pass. Importantly, our pipeline remains entirely unsupervised and self-contained, with no additional training, distillation, or human intervention required. In summary, CLASP harnesses the strengths of deep spectral segmentation—exploiting rich self-supervised features and graph partitioning—while introducing key innovations in automatic cluster selection and direct spectral partitioning. This results in a more principled, efficient, and robust unsupervised segmentation approach that remains true to the one-shot, fully unsupervised paradigm.

\section{Methodology}
\label{sec:method}

\subsection{Small Distilled DINO with Register Tokens}
We leverage a compact variant of the DINO backbone, \texttt{dinov2\_vits14\_reg}~\cite{darcet2024vision}, which integrates register tokens to enrich its feature representations. With just 21 million parameters, this model is ideally suited for real-time applications and environments with limited computational resources. As a Vision Transformer (ViT)-based architecture, DINO generates high-quality patch-level embeddings while also capturing essential global context. Register tokens aggregate global information, improving the stability and structure of learned features, which enhances their utility for clustering and segmentation. In our framework, we compute the cosine similarity between the patch embeddings in the final layer and use these similarities to construct an affinity matrix for spectral clustering. Notably, the model is trained with a cosine contrastive loss, which promotes consistency among embeddings of semantically similar patches. In our method, we extract features from the final layer DINO and do not fuse with any other layer in the network. 

\subsection{CLASP Workflow}
The segmentation workflow is depicted in Figure~\ref{fig:workflow} and Algorithm.~\ref{alg:segmentation}.

\subsubsection{Eigengap Elbow Heuristic}

An image is loaded and resized such that its dimensions are rounded down to integer multiples of 14, matching DINO's $14 \times 14$ patch size. The resized image is then passed through the DINO model, which divides it into non-overlapping patches and extracts a feature vector for each patch that captures both local details and global context.

We then build an affinity matrix \( A \in \mathbb{R}^{n \times n} \) that quantifies the similarity between the feature vectors of the \( n \) image patches. Each element \( A_{ij} \) is defined as 

\begin{equation}
A_{ij} = \frac{\mathbf{f}_i \cdot \mathbf{f}_j}{\|\mathbf{f}_i\| \, \|\mathbf{f}_j\|},
\label{eq:A_calc}
\end{equation}
where \(\mathbf{f}_i\) and \(\mathbf{f}_j\) are the feature vectors for patches \( i \) and \( j \), respectively.

After computing the affinity matrix \( A \), we compute its spectral decomposition. And it is worth noting the choice of using affinity matrix \( A \) instead of Laplacian is intentional. As the affinity matrix \( A \) directly captures pairwise similarities, it preserves the natural clustering geometry of the embeddings without distortion from the degree normalization inherent in the Laplacian transformation. 

    \[
A = Q \Lambda Q^\top,
\]

where \(\Lambda = \operatorname{diag}(\lambda_1, \lambda_2, \ldots, \lambda_n)\) with
\[
\lambda_1 \geq \lambda_2 \geq \cdots \geq \lambda_{n}.
\]
The eigengaps are defined as the differences between consecutive eigenvalues:
\[
\delta_i = \lambda_{i} - \lambda_{i+1},\quad i=1,2,\ldots,n-1.
\]

These eigengaps typically decrease in a nearly monotonic fashion. The eigengap elbow method aims to pinpoint the location in this sequence where the decrease noticeably slows—often referred to as the "elbow point." This point marks the transition from significant improvements in cluster separation (large eigengaps) to diminishing returns (small eigengaps).

\begin{figure}[h]
\begin{center}
\includegraphics[width=0.9\linewidth]{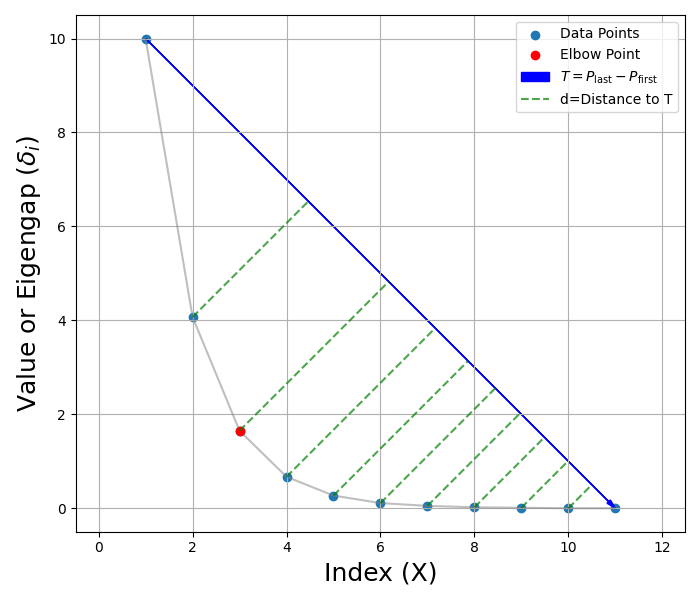}
\end{center}
\caption{This figure illustrates the eigengap heuristic method. The blue dots represent the data points. The solid blue line, labeled 
$T$, connects the first and last points. From each data point, a green dashed line is drawn perpendicular to $T$, with length $d$. The red dot marks the “elbow” point, where 
$d$ is maximized.}
\label{fig:elbow}
\end{figure}

To identify this elbow, we analyze the curvature of the eigengap plot, as illustrated in Fig.~\ref{fig:elbow}. Specifically, we:
\begin{enumerate}
    \item Plot the points \( P(i) = (i, \delta_i) \) for \( i = 1, \ldots, n-1 \), where \(\delta_i\) is the difference between consecutive eigenvalues.
    \item Draw a straight line connecting the first point \( P(1) = (1, \delta_1) \) and the last point \( P(n-1) = (n-1, \delta_{n-1}) \).
    \item Compute the perpendicular distance \( d(i) \) from each point \( P(i) \) to this line.
    \item Identify the index \( i^* \) that maximizes \( d(i) \):
    \[
    i^* = \arg\max_{i} d(i).
    \]
\end{enumerate}
The optimal number of clusters is then defined as:
\[
K_{\text{opt}} = i^* + 1.
\]

Intuitively, if the data naturally forms \(K\) clusters, the first \(K\) eigenvalues will be relatively high and decay sharply, whereas the \((K+1)^\text{th}\) eigenvalue exhibits a marked drop. This distinct eigengap reveals a natural separation between the primary cluster structure and the residual noise or finer details. Overall, this data-driven and intuitive approach provides an interpretable signal for determining the appropriate number of clusters without prior knowledge.

\begin{algorithm}[h]
\small
\caption{CLASP Procedure}
\label{alg:segmentation}
\begin{algorithmic}[1]
\State \textbf{Input:} Image $I$, bandwidth parameter $\beta$
\State $I_{\text{resized}} \gets \text{Resize}(I)$
\State $F \gets \text{DINO}(I_{\text{resized}})$ \Comment{Extract patch feature vectors}
\State $A \gets \text{ComputeAffinity}(F)$ \Comment{Construct affinity matrix (see Eq.~\ref{eq:A_calc})}
\State $[\lambda, Q] \gets \text{EigenDecomposition}(A)$ \
\State $K_{opt} \gets \text{EigengapHeuristic}(\lambda)$ 
\State $S_{\text{best}} \gets -\infty$
\For{$K = \lfloor K_{opt}(1-\beta) \rfloor$ \textbf{to} $\lceil K_{opt}(1+\beta) \rceil$}
    \State $U \gets Q_{(:,1:K)}$ \Comment{Select first $K$ eigenvectors}
    \State $\text{Clusters} \gets \text{SpectralClustering}(U, K)$
    \State $S \gets \text{SilhouetteScore}(\text{Clusters})$
    \If{$S > S_{\text{best}}$}
        \State $S_{\text{best}} \gets S$
        \State $\text{BestClusters} \gets \text{Clusters}$
    \EndIf
\EndFor
\State $M_{\text{patch}} \gets \text{CreatePatchMask}(\text{BestClusters})$
\State $M \gets \text{DenseCRF}(M_{\text{patch}})$ \Comment{Refine segmentation boundaries using DenseCRF}
\State \Return $M$
\end{algorithmic}
\end{algorithm}

\subsubsection{Post-Eigengap Processing and Refinement}

Once the number of clusters \(K_{\text{opt}}\) is estimated using the eigengap heuristic, further refinement is necessary to achieve a robust and detailed segmentation. This stage consists of two major steps: exploration of nearby cluster counts with validation via the silhouette score, and subsequent refinement from patch-level segmentation to a precise pixel-level mask.
\paragraph{Exploration of Cluster Counts:}  
Rather than fixing the cluster count at \(K_{\text{opt}}\), we consider a range of candidate values to account for variability in the data structure. Specifically, we introduce a \emph{bandwidth parameter} \(\beta\) (where \(0 < \beta < 1\)) and define the search interval for \(K\) as
\[
K \in \left[\left\lfloor K_{\text{opt}} (1 - \beta) \right\rfloor, \; \left\lceil K_{\text{opt}} (1 + \beta) \right\rceil \right].
\]
This interval allows us to explore cluster counts slightly below and above \(K_{\text{opt}}\).

For each candidate \(K\) in this range, spectral clustering is performed in the eigenvector space derived from the Laplacian \(L\). The quality of the clustering is then assessed using the silhouette score \(s\). For each patch \(i\), the silhouette score is defined as
\[
s(i) = \frac{b(i) - a(i)}{\max\{a(i), b(i)\}},
\]
where \(a(i)\) is the average dissimilarity (e.g., Euclidean or cosine distance) between patch \(i\) and all other patches within the same cluster, and \(b(i)\) is the minimum average dissimilarity between patch \(i\) and all patches in any other cluster.

The overall silhouette score \(S\) is computed as the mean over all \(n\) patches:
\[
S = \frac{1}{n}\sum_{i=1}^{n} s(i).
\]
The candidate \(K\) that yields the highest \(S\) is then selected, ensuring that the final segmentation demonstrates strong intra-cluster cohesion and clear inter-cluster separation.

\subsubsection{CRF-Based Refinement}

To convert the coarse patch-level segmentation into a refined pixel-level mask, we apply a DenseCRF~\cite{krahenbuehl2011efficient}. Following standard practice, we convert the binary patch mask \(M_{\text{patch}} \in \{0,1\}^{H \times W}\) into a label image and then leverage DenseCRF to integrate spatial and appearance cues, thereby aligning the segmentation boundaries with the object contours.

In summary, the post-eigengap stage bridges the gap between the coarse patch-level segmentation and the final detailed pixel-level mask. By exploring a range of cluster counts, validating with the silhouette score, and refining with a DenseCRF, our approach ensures both high quality segmentation and accurate boundary delineation.





\begin{table*}[t]
\centering
\begin{tabular}{lcccc}
\toprule
 Dataset & \multicolumn{2}{c}{COCO-Stuff} & \multicolumn{2}{c}{ADE20K} \\
\cmidrule(lr){1-1} \cmidrule(lr){2-3} \cmidrule(lr){4-5}
Method         & mIoU  & PixelAcc & mIoU  & PixelAcc \\
\midrule
U2Seg          & 30.2  & 63.9     & --    & --     \\
STEGO          & 28.2  & 56.9     & --    & --     \\
Deep Spectral Segmentation  & 22.0 & 52.1    & 17.8 & 48.1  \\
CLASP (patch)  & 34.4 & 62.4 & 34.7 & 63.5 \\
CLASP (pixel)  & \cellcolor[HTML]{D9EAD3}36.1 & \cellcolor[HTML]{D9EAD3}64.4 & \cellcolor[HTML]{D9EAD3}35.4 & \cellcolor[HTML]{D9EAD3}65.3 \\
\bottomrule
\end{tabular}
\caption{Comparison of segmentation performance across different methods on the COCO-Stuff and ADE20K datasets. We report mean Intersection-over-Union (mIoU) and pixel accuracy (PixelAcc) for each approach. The best-performing results for each dataset are highlighted in green. Our proposed method, \textbf{CLASP}, achieves the highest performance across both datasets, with the \textbf{pixel-based variant} yielding the best overall results.}
\label{tab:performance}
\end{table*}

\section{Experiments}
\label{sec:experiments}

\subsection{Datasets}
We evaluate CLASP on two widely used benchmarks in semantic segmentation. First, we employ the COCO 2017 validation set~\cite{caesar2018coco} with COCO-Stuff annotations, which consists of 5,000 images featuring diverse scenes with varying object counts and complex backgrounds. Second, we assess our method on the ADE20K validation set~\cite{zhou2017scene}, which comprises 2,000 images covering a broad range of indoor and outdoor scenes. 

\subsection{Comparison Methods}
We compare CLASP with several state-of-the-art unsupervised semantic segmentation methods. Specifically, we consider:
\begin{itemize}
    \item \textbf{STEGO}~\cite{hamilton2022unsupervised}, which leverages self-supervised ViT features and distills semantic correspondences into a trainable segmentation network using a contrastive loss to enhance segmentation quality;

    \item \textbf{U2Seg}~\cite{niu2023unsupervised}, which exploits image-level consistency and hierarchical cues to refine segmentation boundaries;
    \item \textbf{Deep Spectral Segmentation}~\cite{melas2022deep}, which integrates deep feature extraction with spectral clustering to uncover latent image structures.
\end{itemize}
For all baseline methods, we adopt the default parameters as specified by the authors.

\subsection{Implementation Details}
In CLASP, we set the bandwidth parameter to \(\beta = 0.5\) to explore candidate cluster counts around the optimal value determined by the eigengap heuristic. For the post-processing stage, we employ a DenseCRF to refine the coarse segmentation masks. Specifically, we configure the DenseCRF with 20 iterations, a ground truth probability of 0.8, and disable uncertain labeling. The Gaussian kernel is parameterized with a spatial standard deviation of 4 and a compatibility weight of 4, while the bilateral kernel is set with a spatial standard deviation of 80, a color (RGB) standard deviation of 13, and a compatibility weight of 10. These settings were selected to strike an effective balance between preserving fine details and enforcing spatial consistency.

We extract patch embeddings from the final layer of DINO, with the embedding dimensions varying according to the input image size. Specifically, we partition the image into patches by performing integer division of the height and width by 14, typically yielding around 2000 patches (e.g., \(40 \times 50\) patches).

\subsection{Evaluation Metrics}
We report standard segmentation metrics, including mean Intersection-over-Union (mIoU) and Pixel Accuracy, to comprehensively assess the segmentation quality and boundary delineation across varied scenes.

\section{Results}
\label{sec:results}

\subsection{Quantitative Evaluation}  
Table~\ref{tab:performance} summarizes the performance on the COCO-Stuff and ADE20K datasets. Notably, the CLASP (pixel) variant achieves an mIoU of 36.1\% and a PixelAcc of 64.4\% on COCO-Stuff, and 35.4\% mIoU with 65.3\% PixelAcc on ADE20K—outperforming U2Seg, STEGO, and Deep Spectral Segmentation by significant margins whilst not requiring any training or further fine-tuning. 

It is important to note that U2Seg, STEGO, and Deep Spectral Segmentation are unsupervised, training data–dependent methods rather than zero-shot approaches. Their performance can suffer when the mask data is biased or low in quality. In contrast, CLASP operates in an end-to-end zero-shot manner, effectively segmenting unseen classes. Moreover, U2Seg tends to under-segment objects, while Deep Spectral Segmentation struggles on heavily textured images, which can limit its effectiveness in complex scenes.

We also observe that the performance of the CLASP (patch) variant—without the DenseCRF refinement—is nearly identical to that of the CLASP (pixel) variant. This implies that for applications where ultra-precise mask boundaries are not critical—such as serving as an initial segmentation pass in real-time video processing pipelines—the DenseCRF step can be safely omitted, leading to a reduction in processing time.

\begin{figure*}[t]
\begin{center}
\includegraphics[width=1\linewidth]{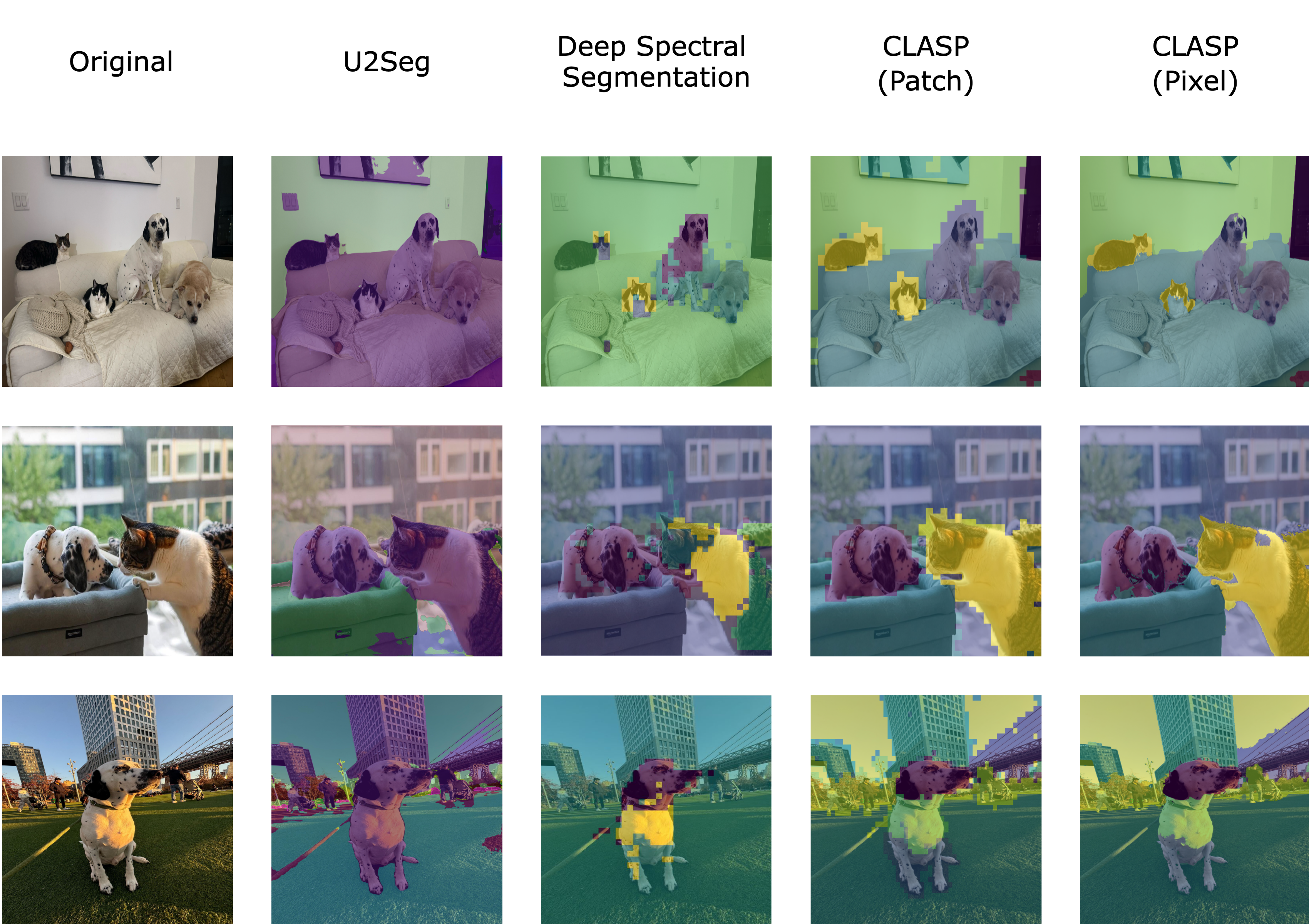}
\end{center}
\caption{Segmentation results on out-of-distribution samples w.r.t. to DINO's training data. Each row shows the original image (left), followed by masks from U2Seg, Deep Spectral Segmentation, and our method CLASP (Patch and Pixel variants). In the top row, CLASP cleanly separates two adjacent animal instances, where other methods merge or miss them. In the second row, CLASP avoids the over-segmentation seen in Deep Spectral while preserving object boundaries. In the final row, although CLASP over-segments the dog, it produces more distinct background segmentation, offering clearer region separation overall. These examples illustrate CLASP’s ability to produce structurally coherent, training-free masks.}
\label{fig:shirley}
\end{figure*}

\subsection{Qualitative Evaluation}
Figure~\ref{fig:shirley} shows qualitative segmentation results on challenging out-of-distribution images. The left column displays the original images, while the subsequent columns compare segmentation masks produced by U2Seg, Deep Spectral Segmentation, and our two CLASP variants (Patch and Pixel). In the \textbf{top row}, CLASP accurately separates adjacent instances of cats and dogs, where competing methods either merge or ambiguously split the objects. In the \textbf{middle row}, our approach avoids the over-segmentation observed in Deep Spectral Segmentation, preserving coherent semantic regions. In the \textbf{bottom row}, although CLASP slightly over-segments a dog, it better differentiates foreground from background elements, resulting in improved overall scene understanding.

\section{Ablation}

We conduct an ablation study to investigate the impact of different parameters and embedding strategies on segmentation performance on the ADE20K dataset without Conditional Random Fields (CRF) for post-processing. Recall comparable performance metrics are $mIoU=34.7\%$ and $PixelAcc=63.5\%$. 

\paragraph{Number of Clusters.}
We set a static number of clusters in our segmentation pipeline. We observed that the highest $mIoU$ was achieved with a fixed cluster number $k$ ranging from 2 to 15, which is $k=9$, resulting in $mIoU=32.1\%$ and $PixelAcc=50.5\%$. It is worth noting that the dataset has on average $8.46$ masks with a long tail. In contrast, the CLASP approach, which dynamically assigns the number of clusters based on the underlying image content, achieved a higher segmentation performance. This result suggests that dynamically determining the appropriate number of clusters, as implemented in CLASP, offers a significant advantage over using a fixed cluster count, likely due to its better adaptation to the image's semantic complexity. 

\paragraph{Patch Embedding Methods.}
To further improve patch-level representations, we experimented with different embedding methods. Notably, we replaced our baseline encoder with the iBOT method~\cite{ibot} for patch-level embedding. iBOT (Image BERT pre-training with Online Tokenizer) is a self-supervised Vision Transformer (ViT) framework that leverages masked image modeling and a teacher-student architecture with online tokenization. Unlike traditional contrastive methods, iBOT learns visual representations by predicting masked patch embeddings, which is particularly beneficial for dense prediction tasks such as segmentation. This allows iBOT to produce more semantically meaningful patch embeddings, potentially enhancing cluster coherence. Using iBOT embeddings, the segmentation $mIoU=33.6\%$ and $PixelAcc=53.3\%$.

\paragraph{Aspect Ratio of the Image}
The aspect ratio of the input image can significantly influence segmentation performance. For iBOT, images are resized to a fixed input size of 225$\times$225 pixels. We evaluated this effect on the dataset using DINOv2, $mIoU=33.2\%$ and $PixelAcc=56.0\%$.

Overall, these results demonstrate that both the choice of cluster number and the underlying patch embedding method play a significant role in segmentation quality.

\section{Conclusion}
\label{sec:conclusion}

We have presented CLASP, a spectral clustering framework for unsupervised semantic segmentation that leverages DINO-based deep features and an adaptive eigengap heuristic to automatically determine the optimal number of clusters. Without any additional training or manual tuning, CLASP achieves competitive performance on the challenging COCO-Stuff dataset, demonstrating the potential of pure spectral methods when combined with modern self-supervised representations.

Beyond its strong performance in image segmentation, CLASP’s simplicity and flexibility open up exciting avenues for broader applications. Its core design—relying solely on graph-based partitioning of deep features—paves the way for efficient and scalable segmentation in diverse domains such as video analysis, audio processing, and text.

\subsection{Future Work}
In future work, we plan to explore alternative strategies for determining the optimal number of clusters. One promising direction is to compare the conventional eigengap heuristic with a log-entropy measure derived from the effective rank of the eigenvalue spectrum. This alternative may capture the overall distribution of eigenvalues more robustly, particularly in scenarios where the spectral decay is gradual, thereby providing a more stable estimate of the number of significant clusters.

Finally, we aim to examine the dispersion statistics of clusters, with a particular focus on identifying background regions, which often exhibit higher variance. A reliable heuristic to distinguish these could further refine segmentation performance, especially in complex scenes where the background may mask subtle foreground details.

These directions are promising as they have the potential to improve the stability and accuracy of unsupervised segmentation and enhance our understanding of spectral clustering dynamics in high-dimensional feature spaces. 

\section{Acknowledgements}
We thank our colleagues for their insightful feedback. This work was supported by Integral Ad Science. We also thank Victor Rambaud for the many meaningful conversations that helped shape this work. Special thanks to Shirley Davila for allowing us to photograph her pets—Petey, Stevie, Minnie, and Annie—who served as models in Fig.~\ref{fig:shirley}.

\end{document}